\documentclass[letterpaper, 10 pt, conference]{ieeeconf}  

\IEEEoverridecommandlockouts                              

\usepackage{graphicx}
\usepackage{amsmath} 
\usepackage{amssymb}  

\usepackage{amsthm}
\usepackage{siunitx}
\usepackage{bm}
\usepackage{xcolor}
\usepackage{float}
\usepackage{algorithm}
\usepackage{algpseudocode}
\usepackage[colorinlistoftodos]{todonotes}
\usepackage{svg}
\usepackage{dblfloatfix}        
\makeatletter
\let\NAT@parse\undefined
\makeatother
\usepackage[colorlinks=true, linkcolor=black, citecolor=black, urlcolor=black]{hyperref}
\usepackage{cite}

\usepackage[normalem]{ulem}

\newcommand{\nvec}{\mathbf} 
\newcommand{\nmat}{\mathbf} 
\newcommand{\transposed}{^T}
\newcommand{\nset}{\mathcal}

\definecolor{tristancolor}{rgb}{0.5,0.5,0.75}

\title{\LARGE \bf
Towards Universal Shared Control in Teleoperation Without Haptic Feedback}

\author{Max Grobbel$^{1}$, Tristan Schneider$^{1}$, Sören Hohmann$^{2}$
\thanks{$^{1}$Max Grobbel and Tristan Schneider are with FZI - Forschungszentrum Informatik, 76135 Karlsruhe, Germany
        {\tt\small grobbel@fzi.de}}%
\thanks{$^{2}$Sören Hohmann is with the Department of Electrical Engineering, Karlsruhe Institute of Technology, Karlsruhe, Germany}
}

\begin{document}

\maketitle
\thispagestyle{empty}
\pagestyle{empty}

\begin{abstract}

Teleoperation with non-haptic VR controllers deprives human operators of critical motion feedback. We address this by embedding a multi-objective optimization problem that converts user input into collision-free UR5e joint trajectories while actively suppressing liquid slosh in a glass. The controller maintains 13 ms average planning latency, confirming real-time performance and motivating the augmentation of this teleoperation approach to further objectives.

\end{abstract}

\section{Introduction}
Teleoperation enables humans to interact with the environment in remote places. Especially inaccessible hazardous environments have been named in research. Further, the lack of human resources, and the need for precision in certain applications require teleoperation systems \cite{sheridan2016}.
Further fields of application include space exploration, handling hazardous materials, surgery and even elder care.
Another upcoming application for teleoperation is the collection of data for imitation learning \cite{wu2024} or preference-based learning \cite{heuvel2025} of robots.

Recent research has shown that teleoperation performance in terms of task completion times is one magnitude worse compared to human performing tasks directly \cite{wu2024}. 
With the perspective by neuroscientist  Daniel Wolpert, that "Movement is the only way you have of affecting the world around you" \cite{wolpert2011}, high performant teleoperation systems might benefit all aforementioned applications.


Most research in teleoperation focuses on displaying information on multiple feedback modalities to the human operator \cite{hauser2025}. A specific focus has been on providing haptic feedback, which leads to significantly higher task performance in tasks with physical interaction \cite{wildenbeest2013}. One downside of such systems is the high cost of up to 40,000 \$ \cite{wu2024}. Alternatively, motion tracking devices as cameras or VR (Virtual Reality) controllers are available at lower cost. 

A common approach to increase task performance in teleoperation is the utilization of shared control algorithms \cite{li2023a}. The overall task is decomposed between the human operator (e.g., commanding the end effector pose) and an automation algorithm (e.g., handling joint-level redundancy or collision avoidance) \cite{grobbel2023}.

This paper contributes a holistic control approach to shared teleoperation by formulating an optimal control problem that contains the robot joints as system states, tracking error and stabilization terms in the objective function, and collision avoidance constraints. This optimal control problem implicitly also solves the inverse kinematics. Initial measurements show that the highly non-convex optimization problem is still solved reliably in real-time and that the parameterization influences the tradeoff between tracking performance and liquid stabilization. 

To showcase the extendability of this approach, the objective function is augmented with a second objective to stabilize liquid in a glass. In this scenario, the human operator is lacking proprioceptive information to stabilize the liquid himself.

\section{Optimal Control Based Telerobotics}
Trajectory planning formulated as an optimal control problem is a widely used approach in robot-manipulator teleoperation \cite{hu2021,rubagotti2019,selvaggio2022}. Leveraging an explicit dynamic model and an objective function, it generates trajectories that both minimize the objective and satisfy the system dynamics. Embedded in a receding-horizon (model-predictive-control) loop, the optimization is resolved at every sampling instant, continuously computing new trajectories.

A key constraint for MPC is meeting real-time deadlines. Advances in processor speed and in numerical solvers have reduced this burden, and robotic implementations now achieve update rates of 1 kHz—even for nonlinear problems—as reported in \cite{faulwassertimm2017}.

In teleoperation, MPC aids the human operator by embedding additional objectives and constraints in the optimization. Collision avoidance, for instance, is enforced by distance constraints \cite{hu2021,rubagotti2019}. Liquid handling poses another case: by modelling the fluid and adding a spill penalty to the cost, the controller yields slosh-free motions \cite{grobbel2023}. Because the human and the automation jointly execute the task, these approaches are referred to as shared control \cite{li2023a,varga2024}.

\section{Teleoperation architecture}
\begin{figure}
    \centering
    \includegraphics[width=0.8\linewidth]{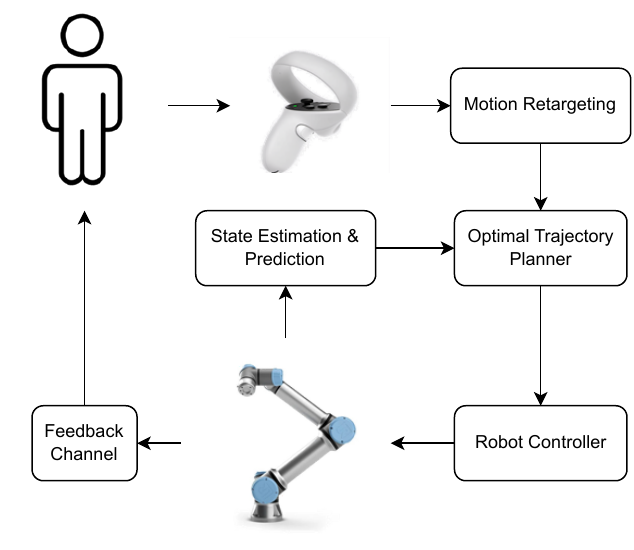}
    \caption{System Architecture of the Shared Control Teleoperation. The central parts of the system are the Motion Retargeting and the Optimal Trajectory Planner. The robot itself interacts with the environment.}
    \label{fig:arcitecture}
\end{figure}

System architectures for teleoperation have been described frequently in literature (e.g. \cite{hokayem2006, siciliano2016}). Typically, those systems are classified by the type of input device. Haptic input devices provide a feedback channel to the human operator, by applying forces and torque to the human operator and come with a higher cost. Those devices usually are some kind of higher dimensional joystick or manipulator. Non-haptic input devices, such as motion capturing devices or cameras, lack the additional feedback channel but come with much lower cost. 

The system under consideration is based on a non-haptic VR-controller. The system architecture is shown in Figure \ref{fig:arcitecture}. The connections between the parts of the system depict information flow. 

The human operator is enabled to communicate goals to the robot through a VR-controller. The process of mapping the pose of the input device onto a desired pose for the robot end effector is referred to as motion retargeting \cite{grobbel2025}. This is the first central part of the system architecture, as task and view point dependent strategies have a huge impact on the overall task performance.

The second central part of the system architecture is the optimal trajectory controller. Including an objective function with multiple goals (tracking and stabilizing a liquid), knowledge about the system dynamics and its constraints, this controller shares the workload with the human operator. The human operator can focus on completing the task in the task space, while the controller solves the inverse kinematics, respects constraints and provides stabilization assistance, which requires local proprioceptive information. 

The trajectory controller generates joint angle trajectories for high frequency tracking controllers, which are embedded on the robot. Feedback is provided by a state estimation, which utilizes sensory information on the robot. The robot itself interacts with the environment. 

The human operator observes the robot and its environment through multiple feedback channels. In the considered setup, the operator has a fixed view point.
An overview of the impact on task performance by modality and quality of the feedback channels is given in \cite{prewett2010}.

\section{Cooperative Optimal Control Trajectory Planning}
An Optimal Control Problem consists of an objective function $J$, system dynamic constraints, and further constraints on the input and state variables. The objective of the control task is given through the objective function, namely reference tracking, liquid stabilization and further stabilization terms. In this section, the modeling of the robot, the selection of objectives and constraints, and the resulting optimization problem for the teleoperation task are described.

\subsection{Robot State Space Modeling}
Previous work has shown that low level controllers of industrial robotic manipulators provide sufficient performance to neglect the dynamics of the physical system in trajectory planning tasks \cite{faulwassertimm2017}. This allows for a linear kinematic system model  
\begin{equation} \label{eq:stateSpaceModelRobot}
    \frac{\mathrm{d}}{\mathrm{d}t}
    {\begin{bmatrix}
    \mathbf{q}\\
    \mathbf{{\dot{q}}}
    \end{bmatrix}}
 = \mathbf{A} \begin{bmatrix}
    \mathbf{q}\\ \mathbf{\dot{q}}
    \end{bmatrix} + \mathbf{B} \mathbf{u}
\end{equation}
with the joint accelerations as control inputs $\mathbf{u} = \mathbf{\ddot{q}}, \mathbf{u} \in \mathbb{R}^{n_q}$ and 
\begin{align}\nonumber
\mathbf{A} = \begin{bmatrix} \mathbf{0} && \mathbf{\mathbb{I}}\\\mathbf{0} && \mathbf{0} \end{bmatrix}
\qquad \mathrm{and} \qquad
\mathbf{B} = \begin{bmatrix} \mathbf{0} \\ \mathbf{\mathbb{I}} \end{bmatrix},
\end{align}
where $\mathbb{I} \in \mathbb{R}^{n_q \times n_q}$ denotes the identity matrix and $\mathbf{0} \in \mathbb{R}^{n_q \times n_q}$. The state vector $\begin{bmatrix}\nvec q\transposed & \dot{\nvec q}\transposed\end{bmatrix}\transposed$ consists of the concatenation of joint angles $\mathbf{q}$ and velocities $\mathbf{\dot{q}}$.

An UR5e robot with six rotational joints is used in this work. The joint vector thus becomes $\begin{bmatrix}
    q_0 & q_1 & q_2 & q_3 & q_4 & q_5
\end{bmatrix}\transposed$. The pose of the end effector is given with the transformation $\mathbf{T}_{0_Me}(\mathbf{q})$ \cite{lynch2017}, which is defined through the forward kinematics using the D-H parameters for the UR5e taken from \cite{UniversalRobotsDH}. In the following, the position of the end effector is denoted with $\mathbf{p}(\mathbf{q})$ and indices are omitted. The orientation of the end effector is given by the rotation matrix $\nmat R(\mathbf{q})$. 

\subsection{Human Movement Prediction}
A prediction of the human reference movement is necessary for a predictive control approach. For this work, a constant velocity along a straight line over the prediction horizon is assumed, leading to a reference of positions $\mathbf{P}_{\mathrm{ref}} = [\mathbf{p}_{\mathrm{ref}, 1}, ... , \mathbf{p}_{\mathrm{ref,N}}]$. To exclude infeasible reference trajectories, predictions extending over the robotic manipulator's reach are clipped. The reference trajectory of orientation $\nmat R_{\mathrm{ref}}$ in the same manner assumes constant angular velocity. More sophisticated approaches like \cite{kille2024}, which also consider the variability in human movement, might be utilized for future experiments.

\subsection{Objective Function}
The main objectives $O_i$ of the trajectory planning task are incorporated in the objective function $J$ of the optimal control problem and defined in the following. In this case, only stage costs $l_i$ are applied. 

\paragraph{$O_1$: Tracking}
The main objective of a teleoperation system is to follow the reference pose given by the operator. The pose can be separated into the position $\mathbf{p}$ and the orientation, given in the form of a rotation matrix $\nmat R$. The prediction of the human motion combined with the motion mapping leads to the reference position trajectory $\mathbf{p}_\mathrm{ref}(t)$ in the manipulator task space.
The squared weighted error is commonly used for position tracking objectives. Thus, the time-dependent stage cost for positional tracking can be defined as 
\begin{equation}
l_{1 \mathrm p}\left(\mathbf{p}, \mathbf{p}_{\mathrm{ref}}\right) = (\mathbf{p} -\mathbf{p}_{\mathrm{ref}})^\mathrm{T} \cdot \mathbf{Q}_p \cdot (\mathbf{p} -\mathbf{p}_{\mathrm{ref}})
\end{equation}
with $\mathbf{Q}_p$ as a diagonal matrix containing the elements of the weight vector $\mathbf{w}_{1p} \in \mathbb{R}^3$, assigning individual weights to the three dimensions in the Euclidean space. Note that the stage cost $l_{1p}$ depends on the position $\mathbf{p}$, which depends on the joint configuration $\mathbf{q}$.


The orientation error is expressed using the rotation matrices $\nmat R$ and $\nmat R_\mathrm{ref}$, which represent the current orientation of the end effector and the desired reference orientation, respectively. Among the various metrics for quantifying orientation errors, we adopt the approach based on the Frobenius norm, which is one of the metrics presented in \cite{huynh2009}. The orientation error cost is formulated as:
\begin{equation}
l_{1 \mathrm o}(\nmat R,\nmat R_\mathrm{ref}) = w_{1 \mathrm o} \cdot  \lVert \mathbb{I} - \nmat R \nmat R_\mathrm{ref}^T \rVert^2_F
\end{equation}
The term $\nmat R \nmat R_\mathrm{ref}^T$ represents the relative rotation between the current and reference orientations. When the orientations are aligned, this product equals the identity matrix $\mathbb{I}$, resulting in zero cost. The Frobenius norm quantifies the deviation from perfect alignment, with the squared norm providing computational advantages by ensuring smooth differentiability and eliminating the need for square root operations during optimization. The orientation error is scaled with the scalar weight $w_{1\mathrm o}$.

\paragraph{$O_{2}$: Objectives for Stability}
The previously defined objective $O_1$ does not include the joint velocities $\mathbf{\dot{q}}$ and acceleration $\mathbf{\ddot{q}}$. For stability reasons it is recommended to define costs for all system variables and inputs. The quadratic cost terms
\begin{align}
    &l_{2 \mathrm a} \left( \mathbf{\dot{q}} \right) = \mathbf{\dot{q}}^T \cdot \mathbf{Q_a} \cdot \mathbf{\dot{q}}\\
    &l_{2 \mathrm b} \left( \mathbf{\ddot{q}} \right) = \mathbf{\ddot{q}}^T \cdot \mathbf{Q_b} \cdot \mathbf{\ddot{q}}
\end{align}
with the diagonal weight matrices $\mathbf{Q_a}$ and $\mathbf{Q_b}$ for velocity $\mathbf{\dot{q}}$ and acceleration $\mathbf{u}$ are added to the objective function. The diagonal entries can be given with the vectors $\mathbf{w}_{2a}$ and $\mathbf{w}_{2b}$.

\paragraph{$O_{3}$: Objective for liquid stabilization} Sloshing of liquid in a container like a glass is often modeled as a pendulum \cite{grobbel2023,biagiotti2023}, where the surface of the liquid is always perpendicular to the rod of the pendulum and the angle of deflection of the pendulum matches the angle of the liquid. The resulting equations of motion can for example be included into the system dynamics (equation (\ref{eq:stateSpaceModelRobot})).

A similar approach focuses on minimizing the lateral accelerations of the glass, viewed in the local reference frame \cite{subburaman2023}. This reduces the complexity of equations to solve in the optimization problem, thus leading to shorter computation times with the downside, that the liquid is not modeled with its own state and differential equation anymore.

The acceleration of the center of the glass is given by twice differentiating the position $\mathbf{p}(\mathbf{q})$ - which is given through the forward kinematics of the manipulator - of the glass with respect to time: 
\begin{align}
    \frac{\mathrm{d}^2}{\mathrm{d}^2t} \mathbf{p}(\mathbf{q}) = \ddot{\mathbf{p}}(\mathbf{q}, \mathbf{\dot{q}}, \mathbf{\ddot{q}}).
\end{align}
Since lateral accelerations in the local frame are minimized, the acceleration in that local frame is calculated by applying the current rotation $\mathbf{R}$. Considering the gravitational acceleration $\mathbf{g} = \begin{bmatrix}
    0 & 0 & 9.81
\end{bmatrix}\transposed$, the local acceleration is calculated with
\begin{align}
   \mathbf{\ddot{p}_{local}} = \mathbf{R} \cdot (\ddot{\mathbf{p}} + \mathbf{g}).
\end{align}

The stabilization of the liquid is achieved by minimizing the lateral accelerations and keeping the vertical acceleration close to the acceleration of gravity. The stage cost is stated as the sum of the squares 
\begin{align}
   l_3(\mathbf{p}, \mathbf{R}) =  (\mathbf{\ddot{p}_{local}} - \mathbf{g})\transposed \cdot \mathbf{Q_3} \cdot (\mathbf{\ddot{p}_{local}} - \mathbf{g})
\end{align}
including the weight diagonal matrix $\mathbf{Q_3} = \mathrm{diag}([w_3, w_3, w_3])$.

The resulting objective function over the prediction horizon $T$ is now given with
\begin{equation} \label{eq:objective}
    J \left( \mathbf{x}, \mathbf{u}, \mathbf{p}_\mathrm{ref}, \nmat R_\mathrm{ref} \right) = \int_0^T l_{1 \mathrm p} + l_{1 \mathrm o} + l_{2 \mathrm a} + l_{2 \mathrm b} + l_3\; \mathrm{d}t,
\end{equation}
where the dependencies on the state trajectories on the right hand side are omitted for readability. 

\subsection{Constraints for Collision Avoidance}
The robot arm geometry is conservatively approximated using $n_\mathrm{s}$ spheres, each characterized by radius $r_i$ and center point $\nvec c_i(\nvec q)$, where $i \in \{1, \ldots, n_\mathrm{s}\}$. Each sphere is assigned to a specific link and moves rigidly with that link according to its forward kinematics. The sphere centers are thus functions of the joint configuration $\nvec q$ through the forward kinematics of their respective links.
Not all sphere pairs require collision avoidance constraints. Spheres on the same link or adjacent links may naturally overlap as part of the conservative approximation, making collision avoidance between them impossible and leading to infeasible optimization problems. Let $\nset I_\mathrm{s} \subset \left\{1,\dots,n_\mathrm{s}\right\}^2$ denote the index set containing all pairs $(i,j)$ of spheres that must remain collision-free.
The collision avoidance constraints are then formulated as:
\begin{equation}
\lVert \nvec c_i(\nvec q) - \nvec c_j(\nvec q) \rVert_2^2 \geq (r_i + r_j)^2, \quad \forall (i,j) \in \nset I_\mathrm{s}
\end{equation}
This formulation ensures that the minimum distance between any two relevant spheres equals or exceeds the sum of their radii, thereby preventing collisions while maintaining computational tractability.

\subsection{Direct Multiple Shooting for Optimal Control}
The system dynamics (\ref{eq:stateSpaceModelRobot}) and the objective function (\ref{eq:objective}) are continuous with respect to time. For implementation, a discretization is necessary. The control input $\mathbf{u}$ is defined as a piecewise constant function. The values $\nvec u_k$ at the discretization nodes are concatenated in the control vector $\mathbf{U}$ with the discrete prediction horizon $N$. The linear system is discretized exactly with this control input using the matrix exponential approach, resulting in a difference equation for $\mathbf{x}_{k+1}$. The integral of the objective function (\ref{eq:objective}) is approximated by a Riemann sum $J_\mathrm{d}$. 
Combining the system model, the reference trajectory and the objective function with path constraints leads to the parameter optimization problem
\begin{align} \label{eq:OCP}
    &\underset{\mathbf{X},\mathbf{U}}{\min}  && J_\mathrm{d} \left( \mathbf{X}, \mathbf{U}, \mathbf{P}_\mathrm{ref} \right) \\
    &\text{s.t.} \nonumber \\
    &&& \mathbf{x}_{k+1} = \mathbf{A}_\mathrm{d} \mathbf{x}_k + \mathbf{B}_\mathrm{d} \mathbf{u}_k  \nonumber\\
    &&& \mathbf{q}_{\min} \leq \mathbf{q}_k \leq \mathbf{q}_{\max} \quad \forall k  \in [1, N] \nonumber\\
    &&& \mathbf{\dot{q}}_{\min} \leq \mathbf{\dot{q}}_k \leq \mathbf{\dot{q}}_{\max} \quad \forall k \in  [1, N] \nonumber\\
    &&& \mathbf{u}_{\min} \leq \mathbf{u}_k \leq \mathbf{u}_{\max} \quad \forall k \in  [0, N-1] \nonumber\\
    &&& \lVert \nvec c_i(\nvec q_k) - \nvec c_j(\nvec q_k) \rVert_2^2 \geq (r_i + r_j)^2, \nonumber \\ 
    &&& \hspace{6em} \forall (i,j) \in \nset I_\mathrm{s}, \, \forall k \in  [1, N]  \nonumber \\
    &&&  \mathbf{x}_0 = \mathbf{x} (k=0) \nonumber.
\end{align}
The chosen minimization with respect to the states $\mathbf{X}$ and the control input $\mathbf{U}$ is called multiple shooting \cite{rawlings2022}.
Even though the system dynamics are linear, the optimal control problem can be assumed to be nonconvex, since the objective function $J$ includes the forward kinematics of the robot manipulator. The resulting state trajectories $\mathbf{X}$ are used as reference inputs to the trajectory tracking controller on the robot. It should be noted that this optimization problem solves the inverse kinematics inherently by incorporating the forward kinematics into the objective function and defining the joint angles in the state vector. 

\section{Experimental Validation}
To analyze the proposed cooperative optimal control approach for teleoperation without haptic feedback, one operator's input with the VR-controller is recorded. This input is used to compare two sets of parameters to analyze the trade-off between tracking performance and liquid stabilization. The experiments demonstrate the real-time capability of the optimization solver.

\begin{figure}[!t]
    \centering
    \includegraphics[width=.95\linewidth]{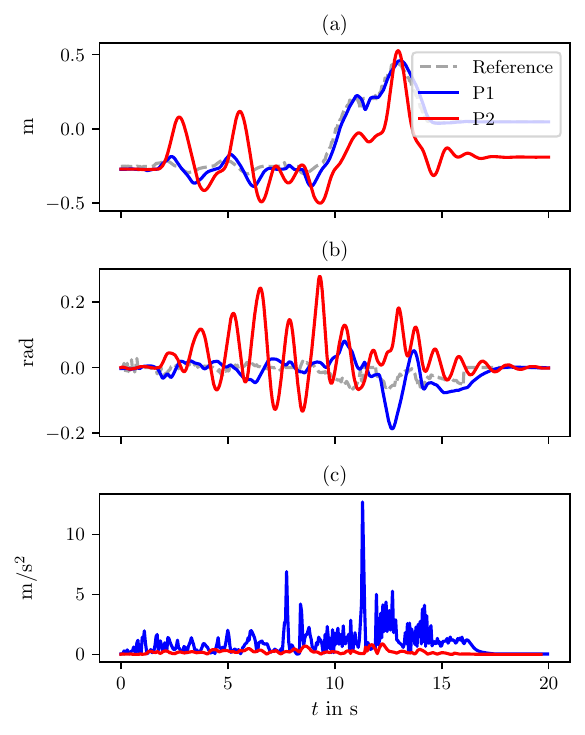}
    \caption{Comparison between tracking and liquid stabilization focus. The same reference trajectory from the input device was used. Plot (a) shows the $y$-position of the end effector and plot (b) shows its roll angle. The third plot (c) shows the local lateral accelerations $(\sqrt{\ddot{\nvec p}_{\mathrm{local},x}^2 + \ddot{\nvec p}_{\mathrm{local},y}^2})$ of the glass, calculated from the planned trajectories.}
    \label{fig:plots}
\end{figure}

\subsection{Hardware and Implementation}

\begin{figure}[ht]
    \centering
    \includegraphics[width=.95\linewidth]{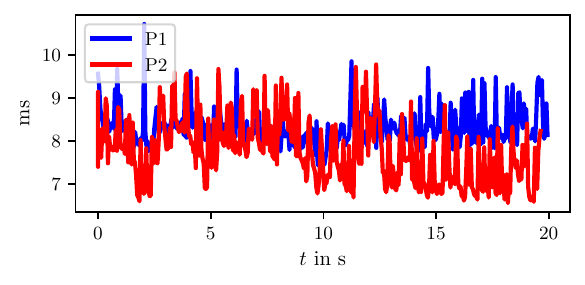}
    \caption{Computation time of the NLP solver.}
    \label{fig:computation_time}
\end{figure}

The experimental setup consists of a Universal Robots UR5e manipulator controlled via a Meta Quest 2 VR controller as the input device. The nonlinear optimization problem given by~\eqref{eq:OCP} is set up in the Python programming language using the framework \emph{CasADi}~\cite{andersson2019} and solved repeatedly with a rate of \SI{20}{\hertz} with the interior point solver \emph{FATROP}~\cite{vanroye2023} on an AMD Ryzen 7950X CPU. The time step size of the discretization is \SI{0.05}{\second} and a prediction horizon of $N = 8$ time steps is used. The geometry of the robot is approximated for collision avoidance using 29 spheres.

The state estimation $\nvec x_0$ is implemented by feeding back the previous optimization solution and interpolating it such that the new $t_0$ corresponds to the current time step. The resulting state trajectories are passed to the ROS2 control trajectory tracking controller.

\begin{table}[h]
\centering
\begin{tabular}{lcc}
\hline
Parameter & P1 (Tracking) & P2 (Anti-slosh) \\
\hline
$\mathbf{w}_{1p}$ & \multicolumn{2}{c}{$[500, 500, 500]$} \\
$w_{1o}$ & \multicolumn{2}{c}{$5$} \\
$\mathbf{w}_{2a}$ & \multicolumn{2}{c}{$[0.1, 0.1, 0.1, 0.1, 0.1, 0.1]$} \\
$\mathbf{w}_{2b}$ & \multicolumn{2}{c}{$[0.02, 0.02, 0.02, 0.02, 0.02, 0.02]$} \\
$w_3$ & $0$ & $10$ \\
\hline
\end{tabular}
\caption{Weights of the objective function for testing the two different control objectives}
\label{tab:parameters}
\end{table}

To study the effects of the liquid stabilization objective, two different sets of parameters are used (\autoref{tab:parameters}). The parameter set P1 corresponds to no liquid stabilization and instead mainly focuses tracking accuracy, while P2 also incorporates liquid stabilization assistance as an objective.

\subsection{Initial Results and Discussion}
Solving the inverse kinematics and tracking of the operator input, collision avoidance, and liquid stabilization can all be included in one optimization problem. With an average of less than 9 ms, the computation time is feasible for real-time applications, with peaks in computation time of up to 11 ms (\autoref{fig:computation_time}). It should be noted, that the reference movement did not target self-collision, in which case the constraint enforcement would have lead to higher computation times. 

\autoref{fig:plots} presents a comparative analysis between the two parameterizations using identical reference trajectories from the VR input device. The results clearly illustrate the multi-objective trade-off inherent in the cooperative control approach:

\textbf{Tracking Performance}: The $y$-position tracking shows that parameterization P1 (tracking-focused) achieves superior reference following compared to P2 (anti-slosh). The tracking deviations are significantly larger for P2, particularly during dynamic maneuvers, demonstrating how the liquid stabilization objective competes with precise end-effector positioning.
The roll angle behavior (middle plot) exhibits similar characteristics, where P2 shows increased deviations from the reference trajectory as the optimization balances orientation tracking against liquid stability requirements, although the tracking weight for orientation was reduced in the whole ablation study to increase degrees of freedom for liquid stabilization.

\textbf{Liquid Stabilization}: During experiments, parameter set P2, specifically tuned for liquid stabilization, prevented liquid spilling, whereas spilling occurred under parameter set P1. As depicted in \autoref{fig:plots}, lateral accelerations planned with P2 were significantly lower - on average \SI{0.18}{\metre\per\second^2} versus \SI{0.94}{\metre\per\second^2} with P1. This behavior was also seen qualitatively, as no liquid was spilled from a full glass, as opposed to P1.

\section{Conclusion}

We demonstrated that a real-time multi-objective optimizer can convert VR-controller inputs into collision-free UR5e manipulator joint trajectories while damping liquid slosh. The planner runs at an average 9 ms latency and is thus real time capable. Experiments with reference trajectories close to constraints should be conducted in the future to test the increase in computational demand. Comparing two weight configurations significantly reduced slosh but sacrificed tracking accuracy. Quantitative slosh metrics showed the qualitatively observed behavior. Although the end-to-end teleoperation system operates reliably, controlled user studies are still required to verify task-level performance gains.




\bibliographystyle{IEEEtran} 
\bibliography{99_references}

\end{document}